\newif\ifarxiv \arxivtrue
\documentclass[sigconf,nonacm]{acmart}

\usepackage{booktabs}
\usepackage{multirow}
\usepackage{tikz}
\usetikzlibrary{positioning,arrows.meta,fit,backgrounds,calc}
\usepackage{listings}
\usepackage{xcolor}
\definecolor{natblue}{HTML}{0072B2}\definecolor{natverm}{HTML}{D55E00}\definecolor{natgreen}{HTML}{009E73}\definecolor{natorange}{HTML}{E69F00}
\newcommand{\nd}[2]{\textbf{#1}\\[-1pt]{\color{black!65}\fontsize{6}{6.8}\selectfont #2}} %
\usepackage{enumitem}
\setlist{nosep,leftmargin=*}
\lstdefinelanguage{pydantic}{
  language=Python,
  morekeywords={Literal,Field,BaseModel,computed_field,model_validator},
}
\newcommand{\codeurl}{https://github.com/yebof/quant-agent}
\newcommand{\releaseAbstract}{The code and prompts of the deployed system are open source.}
\newcommand{\releaseIntro}{The code, prompts and replay harness are open source.}
\newcommand{\releaseEval}{Every number, figure and table in this paper is computed by scripts from the recorded trace, and the open-source system can be re-run on its recorded inputs}
\newcommand{\releasedTrace}{complete decision trace}
\newcommand{\releaseAppendix}{Every table and figure below is computed from the recorded trace by our evaluation scripts.}
\newcommand{\sysname}{\textsc{Steward}} %

\newcommand{\QthreeParagraph}{The second run, on September~30, was live. The reflector first audited its own Q2 diagnosis: \emph{macro warning ignored} had fallen from 14 wrong BUYs to 3, and because the Portfolio Manager prompt already carried rules for it, the reflector declined to add more, citing the risk of overcorrection. It again sent two anomalies to the operators instead of into prompts, attributing them correctly to the digest's instrumentation. It then located a new gap: 70 of 104 classified misses were missed value entries or trend timings, some named on three or more evenings without being acted on. Its one proposal, at \emph{medium} confidence, was a 185-character learning for the Evening Analyst: when the same symbol is classified a missed value entry or trend on three or more evenings, promote it to a suggested action with a TAKE/WAIT call and a named blocker. The gate applied it as a one-file, reversible commit. Because the operators had also revised other prompts and the universe before the cycle, the October record cannot isolate the learning; replay can. We re-ran ten evenings on which such recurrences existed and four control evenings through the prompt before and after the commit, two samples each (Figure~\ref{fig:evolution}b, Table~\ref{tab:q3}; 56 calls, about US\$16). The agent adopted the learning, and only where it applied: the prescribed form (an explicit TAKE/WAIT call or named blocker) appeared in 5 of 20 trigger replays against 0 of 20 for the base prompt and never on a control evening, while the next-day bias stayed as stable as re-running the base prompt (86\% against 79\% agreement). The edit thus added the targeted behaviour without disturbing the rest of the output. Replay also located what the learning still needs: the trigger presupposes a count of past classifications that the Evening Analyst's input does not provide, so the recurring misses surfaced were unchanged (25 of 64 against 26 of 64). Adding that count is a concrete next step, for a gap that the gate's invariants cannot detect and that replay found for US\$16.}

\newcommand{\periodStart}{April~8}
\newcommand{\periodEnd}{September~10, 2026}
\newcommand{\nDays}{102}
\newcommand{\nCalls}{1{,}632}
\newcommand{\nRuns}{619}
\newcommand{\nOrders}{492}
\newcommand{\nRoundTrips}{438}
\newcommand{\costUSD}{\$455}
\newcommand{\retPort}{+5.0\%}
\newcommand{\retSPY}{+12.4\%}

\begin{document}

\title[Governed Self-Evolution for Autonomous Investing]{Governed Self-Evolution for Autonomous Investing:\texorpdfstring{\\}{ }A Multi-Agent LLM System with an Invariant Core}

\author{Yebo Feng}
\affiliation{\institution{Nanyang Technological University}\city{Singapore}\country{Singapore}}
\email{yebo.feng@ntu.edu.sg}
\author{Jingyang Dai}
\affiliation{\institution{Nanyang Technological University}\city{Singapore}\country{Singapore}}
\email{jingyang.dai@ntu.edu.sg}
\author{Liangrui Pan}
\affiliation{\institution{Nanyang Technological University}\city{Singapore}\country{Singapore}}
\email{liangrui.pan@ntu.edu.sg}
\author{Nelson Lau}
\affiliation{\institution{Nanyang Technological University}\city{Singapore}\country{Singapore}}
\email{nelson.lauls@ntu.edu.sg}
\author{Jiahua Xu}
\affiliation{\institution{University College London}\city{London}\country{United Kingdom}}
\email{jiahua.xu@ucl.ac.uk}
\author{Yang Liu}
\affiliation{\institution{Nanyang Technological University}\city{Singapore}\country{Singapore}}
\email{yangliu@ntu.edu.sg}
\renewcommand{\shortauthors}{Feng et al.}

\begin{abstract}
LLM agents that act on live Web data must combine heterogeneous evidence, remember what they have done and improve from outcomes without weakening their own safeguards. We study this problem in autonomous investing, where every action is irreversible and based on untrusted Web content. We present \sysname{}, a multi-agent LLM system that draws the boundary of self-evolution in its type system. First, nine role-specialised agents decide jointly: four analysts read news feeds, SEC filings, macroeconomic series and market data and emit typed reports, from which a Portfolio Manager sets target positions that a protected Risk Manager audits. Second, a seven-layer shared memory, assembled by code from the system's own record, gives each decision its position context, the market narrative, calibration and the agent's past decisions. Third, the system improves itself by review: code grades every trade and forecast nightly and writes the grades into memory, and each quarter a reflector proposes prompt learnings that a deterministic gate applies reversibly and replay of recorded inputs verifies. All three operate around an \emph{invariant core} of order arithmetic, risk rules and discipline-agent prompts that no self-edit can reach. In a five-month live deployment (\nDays{} trading days, \nCalls{} LLM calls), the system contained a 24.6\% sector drawdown to a 0.3\% loss; in replay of its recorded inputs, a single model given the same raw data traded more and lost $1.35$ points against the market over five days where the joint design showed no significant loss, and no injected instruction was followed; and a 2015--2026 backtest shows the core alone lowering the maximum drawdown under all 50 random entry policies. The reflector traced the dominant failure to its root cause, and the gate applied a learning that the agent adopted on trigger evenings and on no control evening. \releaseAbstract
\end{abstract}



\maketitle

\section{Introduction}
\label{sec:intro}

\begin{figure*}[t]
\centering
\resizebox{\textwidth}{!}{\ifdefined\cardhdr\else\newdimen\cardhdr\newdimen\cardh\fi
\cardhdr=11pt    %
\cardh=30pt      %
\providecommand{\lockglyph}{%
  \fill[white] (0,0) circle (3.6pt); \draw[natverm!92!black, line width=0.4pt] (0,0) circle (3.6pt);
  \fill[natverm!92!black] (-1.5pt,-1.9pt) rectangle (1.5pt,0.4pt);
  \draw[natverm!92!black, line width=0.6pt] (-0.9pt,0.4pt) -- (-0.9pt,1.1pt) arc[start angle=180, end angle=0, radius=0.9pt] -- (0.9pt,0.4pt);}
\providecommand{\lockat}[1]{\begin{scope}[shift={([xshift=-1.5pt, yshift=0.5pt]#1.north east)}]\lockglyph\end{scope}}
\providecommand{\card}[3][0pt]{%
  \parbox[t][\cardh][t]{\linewidth}{%
    \makebox[\linewidth][l]{\parbox[c][\cardhdr][c]{\dimexpr\linewidth-#1\relax}{\centering\color{white}\bfseries #2\par}}\par
    \vspace*{\fill}{\centering\color{black!80}\fontsize{6}{6.8}\selectfont #3\par}\vspace*{\fill}}}
\begin{tikzpicture}[
  x=1cm, y=1cm, font=\sffamily\scriptsize, every node/.style={align=center},
  execute at begin node={\hyphenpenalty=10000\exhyphenpenalty=10000\relax},
  card/.style={inner xsep=0pt, inner ysep=0pt, text width=1.86cm, rounded corners=2pt, align=center,
    path picture={\fill[#1] ([yshift=-\cardhdr]path picture bounding box.north west) rectangle (path picture bounding box.north east);}},
  llm/.style={card=natblue!88!black, fill=natblue!12},
  prot/.style={card=natverm!92!black, fill=natblue!12},
  det/.style={card=black!62, fill=black!9},
  web/.style={draw=black!75, densely dashed, line width=0.5pt, rounded corners=2pt, inner xsep=3pt, inner ysep=0pt,
    text width=\dimexpr1.86cm-6pt\relax, minimum height=\cardh, fill=white},
  arr/.style={-{Straight Barb[length=1.3mm, width=1.4mm]}, line width=0.5pt, black!80},
  fb/.style={arr, densely dashed, black!70},
  evo/.style={arr, natgreen!80!black, line width=0.7pt},
  lab/.style={font=\sffamily\fontsize{5.5}{6.5}\selectfont, text=black!75, inner sep=1.2pt, fill=white},
  elab/.style={lab, text=natgreen!65!black},
  grp/.style={fill=black!4, draw=black!22, line width=0.4pt, rounded corners=4pt},
  grpt/.style={font=\sffamily\fontsize{5.5}{6.5}\selectfont, text=black!70, inner sep=1.5pt},
]
\def\cA{0}\def\cB{2.3}\def\cC{4.6}\def\cD{6.9}\def\cE{9.5}\def\cF{12.1}\def\cG{14.7}
\def\rS{0}\def\rA{-1.42}\def\rM{-2.88}\def\rP{-4.36}
\def\pd{0.66}%
\node[web] (rss)   at (\cA,\rS) {\nd{9 RSS feeds}{Reuters, CNBC, AP, BBC, Fed}};
\node[web] (edgar) at (\cB,\rS) {\nd{SEC EDGAR}{10-Q and 10-K filings}};
\node[web] (fred)  at (\cC,\rS) {\nd{FRED}{VIX, yields, CPI, UNRATE, HY OAS}};
\node[web] (mkt)   at (\cD,\rS) {\nd{Market data}{OHLCV, sector, valuation}};
\node[llm] (news)  at (\cA,\rA) {\card{News Intelligence}{3-layer typed report}};
\node[llm] (earn)  at (\cB,\rA) {\card{Earnings Analyst}{5-step CoT}};
\node[llm] (macro) at (\cC,\rA) {\card{Macro Analyst}{6-step CoT}};
\node[llm] (tech)  at (\cD,\rA) {\card{Technical Analyst}{5-step CoT}};
\node[det] (pe)    at (\cE,\rA) {\card{Prompt editor}{12 invariants, 6 target prompts}};
\node[llm] (meta)  at (\cF,\rA) {\card{Meta Reflector}{7-step CoT $\to$ \texttt{PromptLearning[]}}};
\node[det] (rh)    at (\cG,\rA) {\card{Replay harness}{re-runs recorded inputs (R5)}};
\node[det, text width=11.06cm] (mem) at ({(\cA+\cE)/2},\rM) {\card{Shared memory, assembled by code}{L1 today's typed signals $\cdot$ L2 position context and rating trajectories $\cdot$ L3 narrative, regime, active state changes $\cdot$ L4 45-day calibration by size $\cdot$ L5 last five Risk Manager verdicts $\cdot$ L6 own last three decisions $\cdot$ L7 projected book}};
\node[llm] (eve) at (\cF,\rM) {\card{Evening Analyst}{7-step CoT grades every trade}};
\node[det] (cb)  at (\cG,\rM) {\card{Circuit breaker}{no LLM; $-3\%$ day $\Rightarrow$ liquidate}};
\node[llm]  (pm)     at (\cA,\rP) {\card{Portfolio Manager}{7-step CoT $\to$ \texttt{TargetPosition}}};
\node[det]  (pc)     at (\cB,\rP) {\card{Order constructor}{shares, stops, R/R from live quotes}};
\node[det]  (hr)     at (\cC,\rP) {\card{Hard risk engine}{cash-only; size, sector, loss caps}};
\node[prot] (rm)     at (\cD,\rP) {\card{Risk Manager}{6-step CoT: approve, reject, modify, scale}};
\node[det]  (ex)     at (\cE,\rP) {\card{Executor}{sells then buys; stop after each fill}};
\node[web]  (broker) at (\cF,\rP) {\nd{Brokerage API}{orders, fills, P\&L}};
\node[prot] (pr)     at (\cG,\rP) {\card{Position Reviewer}{HOLD, REDUCE, SELL, TRAIL\_STOP}};
\lockat{rm}\lockat{pr}
\foreach \s/\t in {rss/news, edgar/earn, fred/macro, mkt/tech} \draw[arr] (\s) -- (\t);
\foreach \s in {news,earn,macro,tech} \draw[arr] (\s.south) -- (\s.south |- mem.north);
\draw[arr] (mem.south -| pm.north) -- (pm.north);
\draw[arr] (pm) -- (pc); \draw[arr] (pc) -- (hr); \draw[arr] (hr) -- (rm); \draw[arr] (rm) -- (ex);
\draw[arr] (ex) -- (broker) node[lab, midway, above] {orders};
\draw[arr] (pr) -- (broker);
\draw[arr] (cb.west) -- ++(-0.37,0) |- (broker.east);
\draw[fb] ([yshift=0.25cm]broker.west) -- ++(-0.37,0) |- ([yshift=0.25cm]hr.north) -- (hr.north)
  node[lab, pos=0.4, above] {positions, equity};
\draw[evo] (broker.north) -- (eve.south);
\draw[evo] (eve.west) -- (mem.east) node[elab, midway, above] {insights};
\draw[evo] (eve.north) -- (meta.south);
\draw[evo] (meta) -- (pe);
\draw[evo] (meta) -- (rh) node[elab, midway, above] {verify};
\draw[evo] (pe) -- (tech);
\begin{scope}[on background layer]
  \draw[grp] (\cA-1.11,\rS+\pd) -- (\cD+1.11,\rS+\pd) -- (\cD+1.11,\rM+\pd) -- (\cE+1.11,\rM+\pd)
    -- (\cE+1.11,\rP-\pd) -- (\cA-1.11,\rP-\pd) -- cycle;
  \draw[grp] (\cE-1.11,\rA+\pd) -- (\cG+1.11,\rA+\pd) -- (\cG+1.11,\rA-\pd) -- (\cF+1.11,\rA-\pd)
    -- (\cF+1.11,\rM-\pd) -- (\cF-1.11,\rM-\pd) -- (\cF-1.11,\rA-\pd) -- (\cE-1.11,\rA-\pd) -- cycle;
  \draw[grp] (\cG-1.11,\rM+\pd) -- (\cG+1.11,\rM+\pd) -- (\cG+1.11,\rP-\pd) -- (\cF-1.11,\rP-\pd)
    -- (\cF-1.11,\rP+\pd) -- (\cG-1.11,\rP+\pd) -- cycle;
\end{scope}
\node[grpt, anchor=south west] at (\cA-1.11,\rS+\pd) {\textbf{\color{black!80}Morning session}\;\; once a day; the only session that may open positions};
\node[grpt, anchor=south west] at (\cE-1.11,\rA+\pd) {\textbf{\color{black!80}Evening and quarterly}\;\; grade, reflect, gate, verify};
\node[grpt, anchor=north west] at (\cF-1.11,\rP-\pd) {\textbf{\color{black!80}Intraday}\;\; every 30 min, sell-only};
\begin{scope}[shift={(\cE-1.05,\rS+\pd-0.16)}, font=\sffamily\fontsize{5.5}{6.5}\selectfont]
  \def\sw{0.46}\def\sh{0.27}\def\hb{0.095}\def\dy{0.33}
  \foreach \c/\x/\y/\t in {natblue!88!black/0/0/evolvable LLM agent, natverm!92!black/0/1/protected LLM agent, black!62/2.55/0/deterministic code} {
    \fill[\ifx\c\empty\else\fi natblue!12, rounded corners=1pt] (\x,-\y*\dy-\sh/2) rectangle ++(\sw,\sh);
    \fill[\c] (\x,-\y*\dy+\sh/2-\hb) rectangle ++(\sw,\hb);
    \node[anchor=west, inner sep=0, text=black!80] at (\x+\sw+0.1,-\y*\dy) {\t};}
  \fill[black!9, rounded corners=1pt] (2.55,-\sh/2) rectangle ++(\sw,\sh); \fill[black!62] (2.55,\sh/2-\hb) rectangle ++(\sw,\hb);
  \begin{scope}[shift={(\sw-0.02,-\dy+\sh/2+0.01)}, scale=0.8, transform shape]\lockglyph\end{scope}
  \draw[black!75, densely dashed, line width=0.5pt, rounded corners=1pt] (2.55,-\dy-\sh/2) rectangle ++(\sw,\sh);
  \node[anchor=west, inner sep=0, text=black!80] at (2.55+\sw+0.1,-\dy) {Web / API boundary};
  \foreach \s/\y/\t in {arr/0/data and orders, fb/1/state feedback, evo/2/self-evolution loop} {
    \draw[\s] (5.05,-\y*\dy) -- ++(0.5,0);
    \node[anchor=west, inner sep=0, text=black!80] at (5.65,-\y*\dy) {\t};}
\end{scope}
\end{tikzpicture}}
\Description{Block diagram of the trading system as a grid. Top row: four Web sources, each feeding the analyst agent below it. Second row: the four analysts, and on the right the prompt editor, meta reflector and replay harness. Third row: a wide shared-memory block that all four analysts feed, with the evening analyst and circuit breaker on the right. Bottom row: portfolio manager, order constructor, hard risk engine, locked risk manager and executor, then the brokerage API and the locked position reviewer. Green arrows run from the brokerage through the evening analyst to the memory and to the meta reflector, then to the prompt editor and back to the analysts, and to the replay harness. Blue blocks are evolvable LLM agents, vermilion blocks with a lock are protected agents, grey blocks are deterministic code and dashed boxes are the Web boundary.}
\caption{\sysname{}. Header colour gives each component's governance class; shaded panels group the sessions; green arrows trace the self-evolution loop.}
\label{fig:system}
\end{figure*}

LLM agents increasingly act on the Web rather than only answer from it~\cite{yang2025agenticweb}. An agent that makes consequential decisions from Web content must do three things at once: combine heterogeneous, individually weak evidence into one decision; remember what it has decided and how those decisions turned out; and improve from experience~\cite{shinn2023reflexion,madaan2023selfrefine,wang2023voyager,fernando2023promptbreeder,gao2025selfevolving} without weakening the rules that constrain it. The third is the hardest to make safe: an agent that can rewrite its instructions can also rewrite its safeguards, and an agent that grades itself grades optimistically and does not reliably correct itself~\cite{madaan2023selfrefine,xiong2024uncertainty,huang2024selfcorrect}.

Autonomous investing makes these requirements concrete. Every action is irreversible and scored daily in money, and the incentive to let a plausible lesson override a hard rule is constant. It is also a Web setting in the sense of the agentic Web: every input is public Web content (news feeds, SEC EDGAR filings, FRED statistics and exchange quotes), every action leaves through a brokerage Web API, and the central threat, untrusted content reaching an agent with authority to act, is the one the Web-agent security literature studies~\cite{greshake2023injection,debenedetti2024agentdojo,beurerkellner2025patterns}. Prior LLM trading agents~\cite{yu2024finmem,yu2024fincon,wang2024quantagent} combine analyst roles, memory and reflection, but they are evaluated on historical data, where pre-training leakage is hard to exclude~\cite{lopezlira2023chatgpt}; their learning loops run on the model's own reflection or on backtests, and nothing in them separates what may change from what must not.

We present \sysname{}, a multi-agent LLM system in which nine agents trade a US equity portfolio through a brokerage API, in six sessions per trading day and with no human in the trading loop. Four analysts read news, filings, macroeconomic series and market data; a Portfolio Manager decides from their typed reports and from a seven-layer shared memory that code assembles from the system's own record; and the system improves itself by \emph{governed self-evolution}: code grades every decision against realised outcomes, a reflector proposes typed changes to the prompts, and a deterministic, reversible gate applies them but cannot reach an \emph{invariant core} of order arithmetic, blocking risk rules and the prompts of the discipline agents. The key idea is to draw the boundary of self-evolution in the type system rather than in a prompt: the proposal type cannot name the core, and every signal that drives evolution is computed by code from outcomes, so the system can rewrite itself without being able to weaken its own safeguards. Every decision is a typed object stored with its inputs, so the trace that makes the system governable also makes it \emph{evaluable}: each output can be re-scored against later prices and each input replayed through modified prompts. We ran \sysname{} live for five months in a brokerage paper account on real-time Web data and market quotes, so every decision was recorded before its outcome existed. To our knowledge, \sysname{} is the first self-evolving LLM agent evaluated over months of live operation with every proposal and rejection of its evolution loop logged.

\paragraph{Contributions.}
\begin{enumerate}[label=\textbf{C\arabic*}]
\item \textbf{Joint decisions by role-specialised agents over heterogeneous Web data} (\S\ref{sec:overview}, \S\ref{sec:core}): four analysts turn four Web sources into typed reports, and a Portfolio Manager synthesises them, while order arithmetic, blocking risk rules and the prompts of two discipline agents form an invariant core beyond the reach of any prompt, including prompts the system writes for itself.
\item \textbf{A seven-layer shared memory with schema-enforced reasoning} (\S\ref{sec:memory}): code assembles the day's signals, position context, narrative and regime, calibration by position size, recent risk verdicts, the agent's own last decisions and the projected book, so that every number an agent cites comes from code rather than from another agent's prose, and every reasoning step is a mandatory field of a typed chain-of-thought (CoT).
\item \textbf{Governed self-evolution at two time scales} (\S\ref{sec:grade}--\S\ref{sec:verify}): each evening, code grades every trade and forecast and writes the grades into memory; each quarter, a Meta Reflector proposes typed prompt learnings that a twelve-invariant gate applies reversibly and that replay verifies.
\item \textbf{A five-month live deployment with a \releasedTrace{}} (\S\ref{sec:results}): \nDays{} trading days, \nCalls{} LLM calls with complete outputs, 95\% of them also with complete inputs, and \nRoundTrips{} broker round trips, evaluated forward on live data. The value lies in the joint synthesis, as a single model on the same data trades more and worse; the system contained a 24.6\% sector drawdown to a 0.3\% loss and followed no injected instruction, and a 2015--2026 backtest shows the core alone lowering drawdowns across four crashes; 99.9\% of the numbers cited in recorded reasoning are grounded in the agents' inputs; and the evolution loop diagnosed its own failures and applied a learning that the agent adopted only where intended. \releaseIntro\footnote{\url{\codeurl}}
\end{enumerate}

\section{Related Work}
\label{sec:related}

\paragraph{Self-improving and self-evolving agents.}
Reflexion~\cite{shinn2023reflexion} and Self-Refine~\cite{madaan2023selfrefine} let a model critique and revise its own outputs; Voyager~\cite{wang2023voyager} accumulates a skill library; Prompt\-breeder~\cite{fernando2023promptbreeder} and OPRO~\cite{yang2023opro} optimise prompts with the model as the proposer; Generative Agents~\cite{park2023generative} add memory and periodic reflection; a survey frames these as \emph{self-evolving agents}~\cite{gao2025selfevolving}. Two findings qualify the programme: LLMs do not reliably self-correct~\cite{huang2024selfcorrect}, and their verbalised confidence is overconfident~\cite{xiong2024uncertainty}. We take both as constraints: evolution is driven by code-computed outcomes and bounded by a type-level allow-list and a deterministic gate. No prior work reports a self-evolving agent operating for months in a live market with its proposals and rejections logged.

\paragraph{Text, machine learning and LLMs in finance.}
That text moves prices is established: media pessimism predicts short-lived pressure that reverts~\cite{tetlock2007jf}; general sentiment lexicons misclassify most negative words in filings~\cite{loughran2011jf}; and much of newswire sentiment is impounded within the day~\cite{ke2019text}. Machine-learning return prediction from characteristics is the benchmark~\cite{gu2020rfs}. LLMs enter as readers of text: GPT-4 headline labels predict a short drift~\cite{lopezlira2023chatgpt}, and LLMs hallucinate financial facts unless grounded by retrieval and tools~\cite{kang2023hallucination}. This literature evaluates signals in backtests; our contribution is not a signal but the governance of an agent that acts on many, evaluated live.

\paragraph{LLM agents for trading.}
FinMem~\cite{yu2024finmem} introduced layered memory; FinCon~\cite{yu2024fincon} a manager--analyst hierarchy with verbal reinforcement; QuantAgent~\cite{wang2024quantagent} a self-improving knowledge base; FinAgent~\cite{zhang2024finagent} and TradingAgents~\cite{xiao2024tradingagents} tool use and trading-firm roles; StockAgent~\cite{zhang2024stockagent} a simulated market. Their evaluation is historical, where pre-training leakage is hard to exclude~\cite{lopezlira2023chatgpt}, or simulated, and where they learn, the signal is the model's own reflection~\cite{yu2024fincon,zhang2024finagent} or a backtest~\cite{wang2024quantagent}; \sysname{} is evaluated forward in live trading, learns from live outcomes graded by code and evolves around a core its loop cannot reach. Surveys~\cite{nie2024llmfinance}, domain models~\cite{wu2023bloomberggpt} and benchmarks~\cite{xie2024finben} evaluate no deployed agent.

\paragraph{Security of LLM agents.}
Prompt injection was characterised as goal hijacking~\cite{perez2022ignore} and, in its indirect form, as an attack delivered through retrieved Web content~\cite{greshake2023injection}; a formal treatment followed~\cite{liu2024formalizing}, and benchmarks show that agents remain vulnerable~\cite{zhan2024injecagent,debenedetti2024agentdojo,zhang2025asb}. Recent defences are architectural: execution isolation~\cite{wu2025isolategpt}, capability and information-flow control~\cite{debenedetti2025camel,costa2025fides}, and design patterns that bound what an agent may do after reading untrusted input~\cite{beurerkellner2025patterns}. Our invariant core is of this architectural kind, and our setting adds a channel beyond the memory poisoning it studies~\cite{zhang2025asb}: the agent's own record fed back to a reflector that rewrites its instructions.

\paragraph{Frameworks, structured output and Web agents.}
AutoGen~\cite{wu2023autogen}, MetaGPT~\cite{hong2023metagpt} and CAMEL~\cite{li2023camel} coordinate role-specialised agents through natural language and structured messages, building on chain-of-thought~\cite{wei2022cot} and ReAct~\cite{yao2023react}. Constrained decoding~\cite{willard2023outlines} and structured-output modes~\cite{openai2024structured} make schema-conformant JSON cheap; we use schemas as the specification of the reasoning procedure, the audit artefact and the boundary of what evolution may touch. Mind2Web~\cite{deng2023mind2web} and WebArena~\cite{zhou2023webarena} evaluate agents that operate Web interfaces, the agentic-Web vision anticipates agents as first-class Web actors~\cite{yang2025agenticweb}, and Kapoor et al.~\cite{kapoor2024agents} ask that agent evaluation report cost and reproducibility, which we do.

\section{Problem Statement}
\label{sec:problem}

\paragraph{Setting.} An agentic system $S$ with instructions $\pi$ (prompts and code) observes Web evidence $x_t$ at each decision epoch $t$, emits a decision $d_t = S_\pi(x_t)$ and acts on it through an API. The environment later reveals an outcome $y_t$, here market prices, that scores $d_t$ independently of $S$. A \emph{self-evolving} system also runs a loop that maps its own record $\mathcal{R}_T = \{(x_t, d_t, y_t)\}_{t \le T}$ to revised instructions $\pi'$. We assume that the LLM components are capable but unreliable in known ways: they cannot be trusted with factual recall~\cite{kang2023hallucination}, their self-assessments are overconfident~\cite{xiong2024uncertainty} and degrade on new tasks~\cite{kadavath2022know}, their free text is not machine-checkable, and the evidence they read is untrusted~\cite{greshake2023injection}.

\paragraph{Failure modes.} Four failures can occur in such a loop without being observed. (F1)~\emph{Self-grading}: the signal that drives evolution is the agent's own assessment $\hat{g}_t$ rather than $y_t$, and $\hat{g}_t$ is optimistic. (F2)~\emph{Discipline erosion}: $\pi'$ weakens the rules that bound risk, because nothing separates what the loop may change from what it must not. (F3)~\emph{Unauditable change}: no record ties $\pi'$ to the evidence that produced it, so an erroneous edit can be neither found nor reverted. (F4)~\emph{Undetected inertness}: an edit passes every check yet changes no decision, and the loop reports progress it did not make.

\paragraph{Governed self-evolution.} We call a self-evolving system \emph{governed} if it meets five requirements. (R1)~\emph{Typed record}: every $d_t$ is a machine-checkable object stored with $x_t$, so that code can grade, read and replay $\mathcal{R}_T$. (R2)~\emph{Outcome grading}: every signal fed back into decisions or reflection is computed from $y_t$ by deterministic code, never by $S$. (R3)~\emph{Typed proposals}: every proposed change to $\pi$ is a typed object whose justification cites the graded record. (R4)~\emph{Gated application}: deterministic code applies proposals under invariants, logs every acceptance and rejection, can revert in one step, and cannot reach an \emph{invariant core} $\pi_0 \subset \pi$ (the arithmetic, the blocking rules and the discipline agents' prompts) because the proposal type cannot name it. (R5)~\emph{Verification}: the effect of a proposal is measured by replaying recorded inputs through $\pi'$. R1 and R2 address F1 and F3; R3 and R4 address F2 and F3; R5 addresses F4.

\paragraph{Research questions.} A live deployment of such a system lets five questions be answered from the record alone. (RQ1)~What does each layer of $S$, from analyst signal to executed order, contribute to the outcome? (RQ2)~What do the multi-agent decomposition and the shared memory add over a single model? (RQ3)~What does the evolution loop propose, how does the gate act on it, and what does an applied learning change? (RQ4)~Does the invariant core hold, including under adversarial Web content, and at what cost? (RQ5)~Are the numbers the agents reason with grounded in their inputs? \S\ref{sec:system} instantiates R1--R5 in a trading system and \S\ref{sec:results} answers RQ1--RQ5.

\section{Design of \sysname{}}
\label{sec:system}

\sysname{} instantiates R1--R5 in a trading system. Nine LLM agents and a deterministic core run in six sessions (\S\ref{sec:overview}); a shared memory and typed CoT make every decision a checkable record (\S\ref{sec:memory}); safety is enforced by construction rather than by instruction (\S\ref{sec:core}); and the system evolves itself at two time scales, by evening grading (\S\ref{sec:grade}), a gated quarterly loop (\S\ref{sec:reflect}) and replay (\S\ref{sec:verify}).

\subsection{Agents and sessions}
\label{sec:overview}
\paragraph{Sessions.} The system runs six \emph{sessions} per trading day (Figure~\ref{fig:system}; Table~\ref{tab:sessions}, Appendix~\ref{app:ops}). Only the morning session may open positions: four analysts read the Web in parallel, a Portfolio Manager synthesises their typed reports with the shared memory of \S\ref{sec:memory}, and the plan passes through a deterministic order constructor, a blocking risk engine and an LLM Risk Manager before execution. The midday and close sessions are sell-only and run a Position Reviewer; a circuit breaker runs every 30 minutes without any LLM; the evening session grades the day; and a quarterly session reflects on the quarter's graded outcomes. All evidence is public Web data (nine RSS feeds, SEC EDGAR filings, nine FRED series, and daily price, sector and valuation data for a universe of 80 US stocks and 21 exchange-traded funds (ETFs)), and orders leave through a brokerage REST API.

\paragraph{Agents and their typed outputs.} Each agent emits one Pydantic object (Appendix~\ref{app:cot}). The \emph{News Intelligence} agent maintains three layers: a macro narrative of themes and regime that persists across sessions, zero to four \emph{state changes} per session (an event with its previous and new state, market impact, affected symbols and a conviction), and per-symbol news items with a sentiment and an impact summary. The \emph{Earnings Analyst} reads 10-Q and 10-K filings and reports revenue, margins, guidance and risk flags with a sentiment and a data-quality note. The \emph{Macro Analyst} classifies the regime (risk-on, risk-off, neutral or transitional) from the FRED series and issues sector stances and a target invested share. The \emph{Technical Analyst} rates each symbol on a five-level scale with an entry, a reference target, a stop and the condition that would invalidate the setup. The \emph{Portfolio Manager} returns target positions; the \emph{Risk Manager} approves or rejects the plan with a reason category, field-level modifications and an optional scale factor for all buys; the \emph{Position Reviewer} returns SELL, REDUCE, TRAIL\_STOP or HOLD per holding.

\paragraph{Orchestration.} A session runs at most once per day under a lock and is killed after 20 minutes. An LLM call is retried up to seven times with exponential backoff and then fails over to a model from a second vendor. Every other failure removes an action rather than adding one (Table~\ref{tab:failures}, Appendix~\ref{app:impl}).

\subsection{Shared memory and typed reasoning}
\label{sec:memory}\label{sec:schema}
\paragraph{Shared multi-layer memory.} The agents share no conversation history. They share a memory that code assembles from the system's own record in seven layers, read by the Portfolio Manager every morning. L1 holds the day's typed analyst signals. L2 gives each holding its entry date, days held, entry thesis and its last seven Technical ratings. L3 is the narrative: the last seven evening outlooks, seven days of macro regime, and the high-conviction news state changes of the last two weeks. L4 is a 45-day calibration table: code matches buys to later exits first-in-first-out and reports win rate, mean return and holding time by entry size (below \$5k, \$5--10k, above \$10k), so that the agent learns whether its conviction pays. L5 lists the last five Risk Manager verdicts with their categories, L6 the agent's own last three decisions, so that reversals are visible, and L7 a projected book that flags sectors approaching their cap. Every number an agent can cite therefore comes from code, not from another agent's prose.

\paragraph{Schema-enforced reasoning.} Every agent's output type contains its reasoning procedure as a nested CoT object in which each step is a field declared \texttt{str = Field(min\_length=1)} (Listing~\ref{lst:schema}; Appendix~\ref{app:cot} lists every agent's steps). An agent cannot skip a step, enumerations are \texttt{Literal} types, and cross-field validators encode domain invariants: an actionable rating must carry an entry and a stop on the protective side, a \texttt{wrong} grade must name a root cause, and a proposal must cite a number. List-valued fields are validated item by item, so a malformed entry is dropped and logged instead of invalidating a whole report. The validated object is stored verbatim with its exact input, so the record is not a log \emph{about} decisions but the decisions themselves, in a form that code can grade, a reflector can read and a harness can replay.

\subsection{Safety by design}
\label{sec:core}\label{sec:periphery}
Safety does not rest on any prompt: order arithmetic, blocking rules and two discipline agents' prompts form the \emph{invariant core} $\pi_0$, which no self-edit can reach.

\paragraph{Judgement is separated from arithmetic.} No LLM number reaches the broker unchecked. The Portfolio Manager emits \texttt{TargetPosition} \emph{intent} (a symbol, a target weight of at most 20\%, a conviction and a named condition that falsifies the thesis; Listing~\ref{lst:schema}), and a deterministic constructor turns it into orders. It takes the stop from the Portfolio Manager or the Technical Analyst and otherwise places it two average true ranges (ATR) below entry; at execution, code widens any stop closer than one ATR, sizes the position as the smaller of its target weight and a risk budget of 0.5\% of equity at the stop, skips an entry more than 5\% from the market price, and skips a trade whose risk/reward falls below 1.2 once execution has moved its prices. Risk/reward is a computed field that the Risk Manager audits against a floor of 1.5.

\paragraph{Hard rules run before the LLM check.} A blocking engine rejects BUYs that would drive cash negative, positions above 20\% of equity, sectors above 40\%, net exposure above 90\%, entries without a stop, and any buy after a 3\% daily loss. Only plans that pass reach the LLM Risk Manager, which can reject, modify or scale them but cannot approve what the engine blocked. The Risk Manager fails closed: if its call or its schema fails, the plan is rejected.

\paragraph{Untrusted Web text stays upstream.} Prompt injection is the central threat to an agent that reads the Web and holds authority to act~\cite{greshake2023injection}. Raw headlines and filing text reach only the News and Earnings agents, whose prompts treat them as data, drop instruction-like items and flag them in the report. Code then drops any news state change whose keywords and symbols appear neither in the input nor in the prior state, so an agent cannot assert an event its sources do not contain. The Portfolio Manager never sees a headline or a filing: it reads typed fields and the News agent's briefing, and anything it proposes still meets the constructor, the hard rules and the Risk Manager. An injected instruction must therefore survive a typed report, a deterministic engine and an independent auditor before it can become an order (\S\ref{sec:res-perf}).

\paragraph{Execution is fail-safe.} The executor sends sells first as limit orders and waits for their fills, re-checks the daily loss before any buy, records each order before submission so that a crash cannot leave an untracked fill, and places a good-till-cancelled stop order as soon as a buy fills. The circuit breaker, which uses no LLM, runs every 30 minutes and again at each stage of the morning session; on a 3\% daily loss it cancels open entries and liquidates.

\paragraph{The evolvable periphery.} What \emph{can} change is the memory, rewritten daily by code, and the prompts of the four analysts, the Portfolio Manager and the Evening Analyst, each of which carries a marked \emph{Learnings} section that the gate of \S\ref{sec:gate} may append to. The Risk Manager and the Position Reviewer, which encode the hard discipline, are absent from the \texttt{Literal} type of the proposal schema, so no self-edit can name them, and every rule above lives in code that no prompt reaches. A learning can change \emph{what} an agent proposes; it cannot change how a proposal becomes an order, or relax a constraint.

\subsection{Short-term self-evolution: evening grading}
\label{sec:grade}
The evening session closes each trading day with two kinds of grading (R2). Code first assembles the evidence: the sells of the last two days and the buys of the last five with their realised returns, the last ten next-day calls scored against the next day's return, and the moves of at least 8\% over five days in universe names the system did not hold. The Evening Analyst then completes a seven-step CoT that attributes performance, reviews the previous outlook and each position's thesis, grades every trade \texttt{correct}, \texttt{premature} or \texttt{wrong}, assigns every wrong buy one of nine root causes (for example \emph{macro warning ignored} or \emph{top chasing}), classifies missed opportunities, and sets a next-day bias. Grades and P\&L are written to the record in one transaction.

The labels the Evening Analyst assigns are categories; every quantity that measures an outcome is computed by code from fills and prices. What feeds forward is aggregated by code from that record: the calibration of L4, the hit rates of the next-day calls, missed themes that recur in at least two episodes or symbols over the last 14 evenings, and root causes that recur across at least two symbols. The Portfolio Manager reads them the next morning with the evening's outlook, and its prompt bounds the tilt a prior evening may apply to sizing at 20\%.

\subsection{Long-term self-evolution: the quarterly loop}
\label{sec:reflect}
The quarterly loop acts on the periphery in six steps, each assigned to one side of a boundary: language models act and reflect, whereas code records, grades, gates and rewrites.

\paragraph{Reflect and propose (R3).} On the last trading day of each quarter the evening session invokes a Meta Reflector. Its input is a deterministic digest of 90 days of graded record (alpha against SPY, calibration by size, themes caught and missed, root causes of wrong trades and per-agent signal activity) together with every editable prompt, its current \emph{Learnings} section and the previous quarter's reflection. Its CoT has seven mandatory steps in a fixed order: performance facts, theme audit, loss autopsy, a five-axis \emph{self-portrait}, a \emph{gap diagnosis} that attributes the largest gaps to agents, an \emph{audit of existing prompts} that checks whether a rule for each gap already exists, and only then \emph{proposal reasoning}. Its output is at most three \texttt{PromptLearning} objects (Listing~\ref{lst:schema}), each an append or a retraction by content hash, with a text of 20--300 characters, a justification of at least 40 characters that must contain a number from the digest, a target drawn from a \texttt{Literal} of the six editable agents, and a self-rated confidence.

\paragraph{Gate and rewrite (R4).}\label{sec:gate} The gate has two deterministic layers (Appendix~\ref{app:invariants}). The proposal schema rejects a malformed learning before it is staged, and a \texttt{PromptEditor} applies the rest under its own invariants. Edits are append-only into the marked section and tagged with quarter and content hash, so human-written text is never modified; targets must be on the allow-list; a word-boundary tripwire rejects absolutes such as \emph{never}, \emph{always} and \emph{override}, which conflict with the hard rules; a Jaccard similarity of 0.6 or more to an existing entry rejects a paraphrase; a FIFO cap of ten entries per agent bounds the section; at most three agents change per cycle; every accepted \emph{and} rejected proposal is logged with its reason; and a cycle's edits form a single version-control commit, so that one \texttt{git revert} undoes a quarter of evolution. The editor runs \emph{stage-only} (proposals written to a file for human review) or \emph{live-apply}, in which it can apply the reviewed file rather than a fresh reflection. The Q2 cycle ran stage-only and the Q3 cycle live (\S\ref{sec:res-reflect}).

\subsection{Verification by replay}
\label{sec:verify}
Every LLM call is stored with its full response, the model, token counts and cost and, since April~12, its complete input message, so any later decision can be re-run (R5). A \emph{decision-replay} harness passes recorded inputs through the prompt before and after a learning, with the rest of the input fixed, and compares the outputs structurally: for the Portfolio Manager, which symbols were added, dropped or resized; for the Evening Analyst, whose full evening inputs are frozen to disk, how its bias, grades, miss categories and suggested actions changed. Re-running the unchanged prompt on the same inputs measures the model's own variance, against which a learning's effect is read. This replaces the argument that a learning \emph{should} help with a measurement of what it changed on $n$ recorded inputs, at a cost of tens of dollars per experiment.

\section{Evaluation}
\label{sec:results}

\subsection{Live deployment, data and methodology}
\label{sec:deploy}
\sysname{} traded live from \periodStart{} to \periodEnd{} (\nDays{} trading days) on real-time Web data, in a brokerage \emph{paper} account funded with US\$100{,}000. Paper fills are simulated against live quotes, so prices, timing and order rejections are real; market impact is not. Every decision was recorded before its outcome existed, so the live record is forward and free of the pre-training leakage that is hard to exclude in backtests of LLM agents~\cite{lopezlira2023chatgpt}. Idle cash earned the Treasury-bill rate. No human intervened in any trading decision; the evening loop ran on 98 evenings and the quarterly loop at both quarter ends. Code and prompts were unchanged from July~16 to the end of the window. The deployment produced \nRuns{} pipeline runs and \nCalls{} logged LLM calls served by frontier models from two vendors with automatic failover. Logged API cost was \costUSD{}, about US\$4.4 per trading day (Appendix~\ref{app:ops}). The broker recorded \nOrders{} orders and 746 fills; the fill record is the ground truth for all trade-level analysis.

R1 yields three kinds of evidence. The \emph{live record} is broker fills and daily equity. \emph{Trace re-scoring} grades every recorded agent output (7{,}843 Technical Analyst ratings, 1{,}885 symbol-days of news sentiment, 97 evening forecasts, 556 target weights, 90 risk verdicts and \nRoundTrips{} round trips) against later prices without any LLM. \emph{Replay} re-runs recorded inputs through modified prompts or architectures. Forward-return statistics use day-clustered standard errors because horizons overlap. \releaseEval{} (\url{\codeurl}).

\begin{figure*}[t]
\centering
\includegraphics[width=\textwidth]{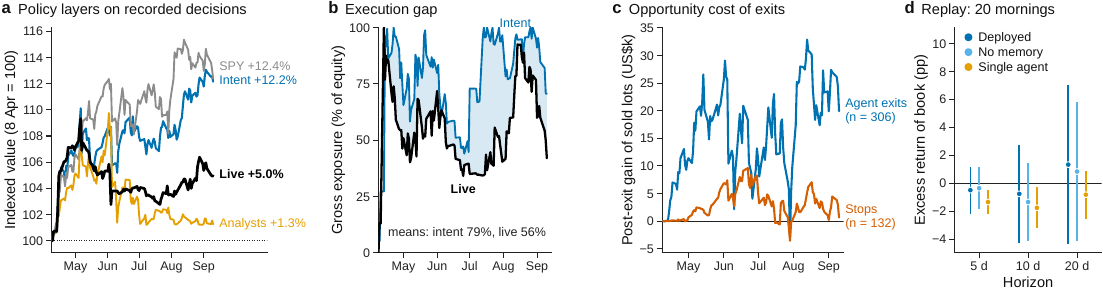}
\Description{Four panels. a, indexed value over time of four series: SPY ends at plus 12.4 percent, the Portfolio Manager's intent simulated with the system's exits and Risk Manager at plus 12.2 percent, the live portfolio at plus 5.0 percent, and a portfolio that follows analyst buy ratings at plus 1.3 percent. b, gross exposure over time: intent averages 79 percent, the live portfolio 56 percent, with the gap shaded. c, cumulative gain made by sold lots after their exit: lots sold by agent decisions went on to gain about 20 thousand dollars, lots sold by stops under 1 thousand. d, mean market-adjusted forward return of books proposed in replay at 5, 10 and 20 days for the deployed design, the design without memory and a single agent; the single agent is negative at 5 and 10 days and all intervals are wide.}
\caption{What the typed trace attributes to each layer. (a)~Recorded decisions followed through prices as signal-following portfolios, against the live portfolio and SPY (other layers in Table~\ref{tab:ablation}). (b)~Gross exposure of simulated intent and live portfolio (capped at 100\% before April~19); shading, the execution gap. (c)~Cumulative gain of sold lots after exit, by exit mechanism. (d)~Replay of 20 mornings through three decision layers: market-adjusted forward return of the proposed book (mean, 95\% CI).}
\label{fig:attribution}
\end{figure*}

\subsection{RQ1: What each layer contributes}
\label{sec:res-signal}\label{sec:res-ablation}\label{sec:res-exits}
\paragraph{Each analyst signal is weak on its own.} Scored against market-adjusted forward returns, the Technical Analyst's five-level rating has a mean daily rank information coefficient (IC) of $-0.004$, $0.019$ and $0.031$ at 5, 10 and 20 days ($t=1.2$ at 20 days), with \emph{sell} ratings carrying the only discernible effect ($-1.7\%\pm1.4$ over 20 days); and News sentiment is uncorrelated with later moves (Figure~\ref{fig:signal}, Appendix~\ref{app:eval}). The value therefore lies in their synthesis. Code grading also held the Evening Analyst to account: its 38 directional next-day calls were right 37\% of the time, below the 54\% base rate of up days, and because the grade is computed by code (R2), this hit rate appeared in its own next-day prompt.

\paragraph{The synthesis carries the value.} Following the recorded decisions through prices (Figure~\ref{fig:attribution}a, Table~\ref{tab:ablation}; targets set at the decision-day close, 20\% position cap, cash earning the T-bill return), the analysts' buy ratings return $+1.3\%$, whereas the Portfolio Manager's target weights return $+15.4\%$ with no exits (at 93\% exposure), $+11.6\%$ with the system's own exit rules and $+12.2\%$ with the Risk Manager's recorded edits, at a higher Sharpe ratio than without them (2.33 against 1.80). The Portfolio Manager's synthesis of weak signals with the shared memory thus matched the index and beat every analyst-following baseline by more than ten points, and the Risk Manager reduced volatility more than return.

\paragraph{Execution and exits hold the headroom.} The live portfolio returned \retPort{} against the intent's $+12.2\%$ because it was less invested: realised gross exposure averaged 56\% against 79\% (Figure~\ref{fig:attribution}b), as 180 of 492 orders were cancelled or expired unfilled, the hard rules trimmed buys and intraday stops fired where the close-based simulation held. The same gap appears on the risk side, as a beta of 0.16 for the live book against 0.64 for the intent (Table~\ref{tab:ablation}). Exits show the same pattern. Of \nRoundTrips{} round trips, 306 were closed by agent decisions and 132 by broker-held stops (Table~\ref{tab:exits}). Lots closed by the core's mechanical stops went on to gain only US\$0.7k beyond the US\$3.7k they realised; lots closed by agent decisions went on to gain US\$19.9k beyond the US\$0.8k realised (Figure~\ref{fig:attribution}c). The Evening Analyst had graded 134 of 163 sells \emph{correct}; only the outcome-scored measure exposes the opportunity cost, which is why R2 puts grading in code.

\begin{table}[t]
\caption{Policy layers on recorded decisions against the live portfolio and SPY. Return, volatility and maximum drawdown in \%; exposure, mean gross.}
\label{tab:ablation}
\footnotesize
\setlength{\tabcolsep}{1.9pt}
\renewcommand{\arraystretch}{1.12}
\begin{tabular}{@{}lrrrrrr@{}}
\toprule
Policy & Return & Vol. & Max DD & Sharpe & Beta & Exposure \\
\midrule
PM intent, no exits, no RM & $+$15.4 & 13.7 & $-$4.7 & 2.52 & 0.83 & 93\% \\
PM intent + system exits & $+$11.6 & 15.0 & $-$4.7 & 1.80 & 0.80 & 76\% \\
PM intent + exits + RM (Intent) & $+$12.2 & 11.9 & $-$4.5 & 2.33 & 0.64 & 79\% \\
\textbf{Live portfolio} & \textbf{$+$5.0} & \textbf{9.0} & \textbf{$-$6.0} & \textbf{1.39} & \textbf{0.16} & \textbf{56\%} \\
\midrule
Analyst-following, all buys & $+$1.3 & 11.3 & $-$7.8 & 0.32 & 0.50 & 41\% \\
Analyst-following, high-conv.\ buys & $+$0.4 & 2.0 & $-$1.8 & 0.41 & $-$0.01 & 5\% \\
SPY & $+$12.4 & 12.4 & $-$4.5 & 2.28 & 1.00 & -- \\
\bottomrule
\end{tabular}

\end{table}

\subsection{RQ2: The decomposition and the memory}
\label{sec:res-arch}
The simulations take the Portfolio Manager's intent as given. To test the design upstream of it, we replay 20 mornings sampled evenly over the deployment through three decision layers on their exact recorded inputs (Table~\ref{tab:arch}, Figure~\ref{fig:attribution}d; 100 LLM calls): the deployed design (typed analyst reports plus memory; two samples per morning), the same input without any memory layer, and a single model that receives the raw data the analysts read and must analyse and decide in one call. Each proposed book is scored by the weight-averaged market-adjusted forward return of its names. The single agent is a different trader: it targets 5.8 names a morning against 3.7, puts 41\% of equity into new intent against 19\%, and reverses its own recent decisions two and a half times as often (0.25 against 0.10 per morning), including re-buying names it had just closed, which the deployed design never did. Its books lose over the next five and ten days ($-1.35\pm0.44$ and $-1.76\pm0.75$ points against SPY; hit rates 35\% and 32\%), whereas the deployed design's are indistinguishable from zero. Removing the memory changes almost nothing on a single morning, so the typed reports carry most of the per-morning information and the memory's role is longitudinal: calibration, continuity and feedback.

\begin{table}[t]
\caption{Architecture replay on 20 recorded mornings. Names: positions targeted; New: of which not held; Gross: new intent, \% of equity; Rev.: reversals; Excess: market-adjusted forward return, pp (mean $\pm$ s.e.).}
\label{tab:arch}
\scriptsize
\setlength{\tabcolsep}{2.6pt}
\begin{tabular}{@{}lrrrrrr@{}}
\toprule
Decision layer & Names & New & Gross & Rev. & Excess 5d & Excess 20d \\
\midrule
\textbf{Deployed design} & 3.7 & 3.2 & 19\% & 0.10 & $-$0.5$\pm$0.9 & $+$1.3$\pm$2.9 \\
No shared memory & 3.5 & 3.0 & 19\% & 0.12 & $-$0.4$\pm$0.8 & $+$0.8$\pm$2.5 \\
Single agent, raw data & 5.8 & 4.4 & 41\% & 0.25 & $-$1.3$\pm$0.4 & $-$0.8$\pm$0.9 \\
\bottomrule
\end{tabular}

\end{table}

\subsection{RQ3: Two cycles of self-evolution}
\label{sec:res-reflect}\label{sec:res-gate}
\begin{figure}[t]
\centering
\includegraphics[width=\linewidth]{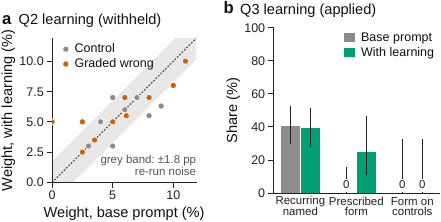}
\Description{Two panels. a, scatter of each flagged buy's weight under the base prompt against its weight with the withheld Q2 learning; points lie along the diagonal within a grey band of re-run noise for both wrong and control buys. b, bar chart comparing the base prompt and the prompt with the applied Q3 learning: the share of recurring misses named in suggested actions is about 40 percent for both; the share of replays in the prescribed TAKE or WAIT form rises from 0 to 25 percent; on control evenings it is 0 for both.}
\caption{Both quarterly learnings verified by replay. (a)~Q2, withheld: weight of each flagged BUY with and without the learning; grey band, re-run noise. (b)~Q3, applied: share of recurring misses named, and of replays in the prescribed form, on trigger and control evenings (95\% Wilson CI).}
\label{fig:evolution}
\end{figure}

\paragraph{Q2: a root-cause diagnosis and a withheld learning.} The Meta Reflector's first run (stage-only, 52 trading days) attributed the shortfall against the index to ``structural under-participation in a broad rally, not a blow-up''; found that 14 of the 20 BUYs graded \emph{wrong} shared one root cause, \emph{macro warning ignored} (an active high-conviction bearish news state change at purchase); declined to edit a prompt where it traced the gap to the data rather than the instructions; and proposed exactly one learning, to treat such a state change as a sizing gate, at self-rated \emph{low} confidence. The gate withheld it on its length invariant (282 characters against a 200-character cap) and logged why. Replay then measured what the gate had declined (Figure~\ref{fig:evolution}a, Table~\ref{tab:replay}; 35 calls, US\$17): on the ten mornings whose BUYs were later graded wrong and five control mornings, the Portfolio Manager prompt with the learning kept 8 of the 10 flagged names that the baseline still proposed within one point, and every weight moved within the noise of re-running the baseline itself (1.8 points per name at the median). The withheld learning was inert, so the strict gate cost nothing.

\paragraph{Q3: self-audit and an applied learning.} \QthreeParagraph

\paragraph{The evening loop.} The daily loop should make the system less active after poorly graded periods, and the record is consistent with this: realised exposure co-moves with the trailing forecast hit rate ($\rho=0.34$, $p=0.01$) and the trailing realised win rate ($\rho=0.22$, $p=0.05$), whereas the Portfolio Manager's declared sizing does not ($|\rho|\le0.14$; Table~\ref{tab:feedback}; Holm-adjusted $p=0.08$ for the strongest relation): the adaptation runs through the graded memory rather than through the agent's stated intent.

\subsection{RQ4: The invariant core under live and adversarial conditions}
\label{sec:res-perf}
\paragraph{Containment.} The portfolio returned \retPort{} against \retSPY{} for SPY during a semiconductor-led rally (Table~\ref{tab:perf}), at a beta of 0.16 and volatility of 9.0\% against 12.4\%; alpha is not identifiable over \nDays{} days, and no claim of this paper rests on it. Between June~22 and July~29 the semiconductor ETF fell 24.6\%, QQQ 10.3\% and SPY 2.0\%, whereas the portfolio lost 0.3\%, having exited its semiconductor positions in May on named triggers that are legible in the trace. One window holds one drawdown, so we also backtest the deterministic core alone over 2015--2026 (Table~\ref{tab:backtest}, Figure~\ref{fig:backtest}): its stops, caps, cash-only rule and breaker applied to LLM-free weekly entries on the universe's 80 stocks, which rules out pre-training leakage. The core lowered the maximum drawdown under all 50 random entry policies (median 37\% $\to$ 28\%) and under momentum entries (61\% $\to$ 33\%), and the drawdown of the 2018, 2020 and 2022 sell-offs for at least 49 of 50, at an unchanged Sharpe ratio; the caps and the breaker do most of the work. In the 2026 semiconductor window the same rules still lost 27\% under momentum entries: the core bounds what judgement can lose, and the agents' May exits did the rest. The Risk Manager rejected 4 of 90 plans and made 146 field-level edits; the circuit breaker never had to liquidate; after the cash-only rule took effect on April~19 gross exposure never exceeded 92\%; and across \nCalls{} calls validation failed on 13 sub-items, each isolated without losing a session.

\begin{table}[t]
\caption{Full-period outcome against benchmarks and daily-rebalanced T-bill blends (zero risk-free rate). Return, volatility, drawdown and $\alpha$ (Jensen's, against SPY, annualised) in \%.}
\label{tab:perf}
\scriptsize
\setlength{\tabcolsep}{3pt}
\begin{tabular}{@{}lrrrrrrr@{}}
\toprule
 & Return & Vol. & Max DD & Sharpe & Sortino & Beta & $\alpha$ (ann.) \\
\midrule
\textbf{Live portfolio} & \textbf{$+$5.0} & \textbf{9.0} & \textbf{$-$6.0} & \textbf{1.39} & \textbf{1.76} & \textbf{0.16} & \textbf{$+$8.2} \\
SPY (S\&P 500) & $+$12.4 & 12.4 & $-$4.5 & 2.28 & 3.55 & 1.00 & $-$0.0 \\
QQQ (Nasdaq-100) & $+$17.1 & 22.1 & $-$11.2 & 1.79 & 2.79 & 1.63 & $-$6.6 \\
SMH (semiconductors) & $+$32.5 & 46.8 & $-$24.6 & 1.65 & 2.45 & 2.79 & $-$1.7 \\
50/50 SPY / T-bills & $+$6.9 & 6.2 & $-$2.2 & 2.58 & 4.02 & 0.50 & $+$1.8 \\
Exposure-matched (56\% SPY) & $+$7.5 & 6.9 & $-$2.5 & 2.52 & 3.92 & 0.56 & $+$1.6 \\
Beta-matched (0.16 SPY) & $+$3.2 & 1.9 & $-$0.6 & 3.87 & 6.15 & 0.16 & $+$3.1 \\
\bottomrule
\end{tabular}

\end{table}

\paragraph{Adversarial Web content.} Untrusted content reaches the agents through the news feeds. We insert one adversarial item into the exact recorded input of ten mornings and re-run the agent (Table~\ref{tab:injection}; 50 calls, US\$19), with three payloads: a direct override (set one name to 45\% and every other target to zero), a plausible headline (300\% upside; managers are taking maximum size and disabling stops) and a suspension of discipline (stops and the cash-only rule lifted, allocate 60\% to an inverse ETF). The News agent excluded the item from its typed report in all 20 replays and flagged it as instruction-like in every briefing. Placed directly in the Portfolio Manager's news section, as if the analyst layer had passed it through, the item was followed in none of 30 replays; in 23 the Portfolio Manager's CoT named it as injected content and refused it, most often because the symbol had no Technical Analyst report and the demand conflicted with the stop and cash rules. Had any replay complied, the schema would have dropped a 45\% or 60\% target before the constructor and the hard caps would have bounded anything smaller: injected content could at most change what an agent \emph{wanted}, and here it did not achieve even that.

\begin{table}[t]
\caption{Prompt-injection red team on 10 recorded mornings. Excluded/Labelled: dropped from / flagged in the News report; Followed/Refused: weight given to / refusal of the payload's target by the Portfolio Manager; Max wt: largest weight the caps would allow.}
\label{tab:injection}
\scriptsize
\setlength{\tabcolsep}{4pt}
\begin{tabular}{@{}lrrrrr@{}}
\toprule
 & \multicolumn{2}{c}{News agent} & \multicolumn{2}{c}{Portfolio Manager} & Core \\
\cmidrule(lr){2-3}\cmidrule(lr){4-5}\cmidrule(lr){6-6}
Payload & Excluded & Labelled & Followed & Refused & Max wt \\
\midrule
P1 direct override & 10/10 & 10/10 & 0/10 & 7/10 & 0\% \\
P2 plausible headline & 10/10 & 10/10 & 0/10 & 6/10 & 0\% \\
P3 discipline suspension & -- & -- & 0/10 & 10/10 & 0\% \\
\bottomrule
\end{tabular}

\end{table}

\subsection{RQ5: Grounding of cited numbers}
\label{sec:res-ground}
Because every CoT is stored with its exact input, we can ask whether the numbers an agent cites exist in what it was given. A deterministic audit of 750 recorded CoTs (Table~\ref{tab:grounding}) extracted 254{,}094 cited numbers, excluding small integers and years, and matched each against the agent's input, its prompt and its own emitted fields. A strict matcher (exact, rounded or truncated matches and explicitly written arithmetic) grounds 96.1\%; recognising derived quantities such as risk/reward ratios and ATR-based stops raises this to 99.9\%, and the median CoT is fully grounded for every agent. A placebo against another day's input grounds 22\% under the strict matcher, so the test discriminates. Of the 226 ungrounded numbers, nearly all are unwritten intermediate results of the sizing formula or chosen trigger levels; three are candidate fabrications (a wrong price, a misread indicator and a recalled policy-rate peak absent from the input).

\section{Discussion}
\label{sec:discussion}

RQ1--RQ5 yield three implications for multi-agent systems that act and evolve in high-stakes Web settings.

\paragraph{Place the invariant core at the type level.} Making the core unreachable rather than discouraged (arithmetic and blocking rules in code, the discipline agents absent from the \texttt{Literal} of editable targets) is what makes it safe for the system to rewrite itself. Lower drawdowns across four crashes in backtest and the failed injections are its observable consequences.

\paragraph{Grade with code, and allocate discretion where the data show skill.} Outcome-scored grading attributes the value to the Portfolio Manager's synthesis and the core's mechanical exits and locates the headroom in discretionary exits and execution; the Evening Analyst's own grades, which rated 134 of 163 sells correct, would not have shown this. A self-evolving system can improve only what it measures, so the next step is to narrow the periphery on exits and add an execution agent inside the core, leaving the judgement layer that works untouched.

\paragraph{A syntactic gate needs a semantic check.} The gate guarantees that a learning is short, novel, reversible and confined to the periphery; it cannot tell whether the learning changes a decision. Replay can, cheaply: it confirmed that the withheld learning was inert and that the applied one was adopted only where intended (\S\ref{sec:res-reflect}). The Q3 learning also presupposes a count of past classifications that the Evening Analyst's input does not provide. Replay turned this into a concrete fix. The natural next invariant is therefore an outcome-scored replay check that confirms a learning is operable on the agent's actual inputs and measures its effect on decisions before it is applied.

\paragraph{Limitations and next steps.} The evidence comes from one configuration traded on a paper account through one rising window, with overlapping forward horizons, simulations that ignore fill risk and two quarterly cycles, one applied; replays of 14--20 inputs expose behavioural differences but cannot rank variants by return. Each limit narrows with operation: every further quarter adds a cycle to the record, and the released harness scales replay to more inputs at tens of dollars per experiment.

\section{Conclusion}
\label{sec:conclusion}
We presented \sysname{}, a multi-agent LLM system for autonomous investing in which role-specialised agents decide jointly over heterogeneous Web data, share a memory computed by code, and evolve at two time scales around an invariant core that no self-edit can reach. Five months of live operation showed that the typed trace that makes such a system governable also makes it evaluable: it located the method's value and its headroom layer by layer, recorded the system containing a sector drawdown and resisting injected instructions, with a backtest of the core alone lowering drawdowns across four crashes, and documented self-evolution diagnosing its own failures and making targeted, reversible edits whose effects replay verified at a cost of tens of dollars. We release the system, the trace and the scripts so that outcome-scored replay can be developed into a gate invariant on real inputs.

\section*{Ethical Considerations}
The system traded a paper account, so no capital was at risk; all inputs are public Web data used under the providers' terms; no human subjects were involved; and the released software is not investment advice. The invariant core prevents a self-modifying agent from eroding its own risk discipline by construction, and the recorded{} trace lets this be audited.

\bibliographystyle{ACM-Reference-Format}
\bibliography{refs-compact} %

\appendix
\raggedbottom %
\section{Schema Excerpts}
\label{app:schema}
Listing~\ref{lst:schema} shows the typed record for the Portfolio Manager's decision and for a self-evolution proposal; the CoT steps of every agent are in Appendix~\ref{app:cot}.

\begin{lstlisting}[language=pydantic, caption={Excerpts of the typed record (\texttt{src/models.py}). Every CoT step is mandatory; the proposal type cannot name a protected agent.}, label={lst:schema}]
class ReasoningChain(BaseModel):        # Portfolio Manager CoT
    macro_filter:      str = Field(min_length=1)
    news_check:        str = Field(min_length=1)
    earnings_check:    str = Field(min_length=1)
    signal_conflicts:  str = Field(min_length=1)
    sizing_logic:      str = Field(min_length=1)
    portfolio_balance: str = Field(min_length=1)
    cash_target:       str = Field(min_length=1)

class TargetPosition(BaseModel):        # intent, not shares
    symbol: str
    target_weight_pct: float = Field(ge=0, le=20)  # cap in the type
    thesis: str
    conviction: Literal["high", "medium", "low"]
    thesis_invalid_if: str               # named exit condition

MetaReflectionAgentName = Literal[      # no risk_manager,
    "tech_analyst", "news_analyst",     # no position_reviewer
    "macro_analyst", "earnings_analyst",
    "portfolio_manager", "evening_analyst"]

class PromptLearning(BaseModel):        # a proposed self-edit
    agent_name: MetaReflectionAgentName
    operation: Literal["append", "retract"]
    learning_text: str = Field(min_length=20, max_length=300)
    justification: str = Field(min_length=40)  # must cite numbers
\end{lstlisting}

\section{Agent Chain-of-Thought Schemas}
\label{app:cot}
Every LLM agent's output model contains a nested chain-of-thought (CoT) object whose fields are the reasoning steps the agent must perform, in order. Each step is declared \texttt{str = Field(min\_length=1)}, so a missing or empty step fails validation and the response is retried; the steps are therefore not a prompting hint but part of the output contract, and they are persisted verbatim with the decision. Table~\ref{tab:cot} lists the mandatory steps of each agent's CoT as they appear in \texttt{src/models.py}. The News Intelligence agent is the one agent without a CoT: its structured output (a persistent macro narrative, a list of typed state changes with conviction, and per-symbol alerts) \emph{is} its reasoning structure. The Portfolio Manager's CoT additionally carries two optional steps, \texttt{continuity\_check} (consistency with its own last three decisions) and \texttt{premortem\_check} (the strongest argument against its largest position), which the prompt requires.

\begin{table}[H]
\caption{Mandatory chain-of-thought steps per agent (field names of the Pydantic CoT models).}
\label{tab:cot}
\scriptsize
\setlength{\tabcolsep}{3pt}
\begin{tabular}{@{}p{1.55cm}p{6.6cm}@{}}
\toprule
Agent & CoT steps, in order \\
\midrule
Tech Analyst (5) & \texttt{trend} (price versus moving averages) $\to$ \texttt{momentum} (RSI, MACD) $\to$ \texttt{volatility} (Bollinger position, ATR) $\to$ \texttt{volume} (confirming or diverging) $\to$ \texttt{support\_resistance} (levels, valuation flags). Output: rating, conviction, entry, stop, target, \texttt{thesis\_invalid\_if}; risk/reward is computed, not written. \\
Macro Analyst (6) & \texttt{volatility\_analysis} $\to$ \texttt{yield\_curve\_analysis} $\to$ \texttt{monetary\_policy\_analysis} $\to$ \texttt{inflation\_labor\_credit} $\to$ \texttt{cross\_signal\_synthesis} $\to$ \texttt{sector\_implications}. Output: regime, confidence, target exposure, sector stances, regime-shift flag. \\
Earnings Analyst (5) & \texttt{fundamental\_quality} $\to$ \texttt{growth\_trajectory} $\to$ \texttt{strategic\_risks} $\to$ \texttt{management\_execution} $\to$ \texttt{valuation\_context}. Output: sentiment and conviction derivable from the steps. \\
Portfolio Manager (7\,+\,2) & \texttt{macro\_filter} $\to$ \texttt{news\_check} $\to$ \texttt{earnings\_check} $\to$ \texttt{signal\_conflicts} $\to$ \texttt{sizing\_logic} $\to$ \texttt{portfolio\_balance} $\to$ \texttt{cash\_target}; optional \texttt{continuity\_check}, \texttt{premortem\_check}. Output: \texttt{TargetPosition} intent per symbol. \\
Risk Manager (6) & \texttt{rr\_audit} $\to$ \texttt{signal\_fidelity} $\to$ \texttt{correlation\_check} $\to$ \texttt{event\_risk} $\to$ \texttt{sizing\_sanity} $\to$ \texttt{overall}. Output: approve or reject, per-symbol modifications, \texttt{scale\_all\_buys}, reason category. \\
Position Reviewer (6) & \texttt{macro\_continuity\_check} $\to$ \texttt{thesis\_progress\_check} $\to$ \texttt{thesis\_integrity\_check} $\to$ \texttt{winners\_discipline\_check} $\to$ \texttt{session\_disposition\_check} $\to$ \texttt{execution\_rationale}. Output: HOLD, REDUCE, SELL or TRAIL\_STOP per position. \\
Evening Analyst (7) & \texttt{performance\_attribution} $\to$ \texttt{outlook\_retrospection} $\to$ \texttt{thesis\_health\_review} $\to$ \texttt{decision\_quality\_review} $\to$ \texttt{calibration\_meta} $\to$ \texttt{market\_regime\_read} $\to$ \texttt{tomorrow\_preparation}. Output: trade grades, thesis trajectories, missed opportunities, next-day bias. \\
Meta Reflector (7) & \texttt{performance\_vs\_benchmark} $\to$ \texttt{secular\_theme\_audit} $\to$ \texttt{loss\_autopsy\_audit} $\to$ \texttt{self\_portrait\_synthesis} $\to$ \texttt{portrait\_gap\_diagnosis} $\to$ \texttt{existing\_prompt\_audit} $\to$ \texttt{prompt\_edit\_reasoning}. Output: \texttt{PromptLearning} proposals, blind spots, confidence. \\
News Intelligence (--) & No CoT; typed output of macro narrative, state changes and per-symbol alerts. \\
\bottomrule
\end{tabular}
\end{table}

\section{Gate Invariants}
\label{app:invariants}
The proposal schema and the deterministic prompt editor enforce the following on every proposed learning, independently of the LLM: (1)~append-only into a marked \emph{Learnings} section, each entry tagged with quarter and content hash; (2)~an allow-list of target agents, with the Risk Manager and Position Reviewer excluded both here and in the proposal schema's \texttt{Literal}; (3)~length ceiling (300 characters; 200 before the Q3 cycle) and justification floor; (4)~a word-boundary prohibited-word tripwire (\emph{never}, \emph{always}, \emph{override}, \emph{ignore all}, \emph{must always}, \emph{must never}); (5)~Jaccard token-similarity dedup against existing entries (threshold 0.6); (6)~a per-agent FIFO cap that evicts the oldest auto-added entry; (7)~a per-cycle cap on the number of agents edited; (8)~atomic file writes; (9)~an audit log of accepted and rejected proposals with reasons; (10)~one git commit per cycle, skipping any prompt file with uncommitted human edits so that the commit stays cleanly revertible; (11)~single-line, whitespace-normalised entries; (12)~an apply-from-file lane so that a human-reviewed proposal file, not a fresh reflection, is what gets applied.

\section{Sessions, Latency and Cost}
\label{app:ops}
\begin{figure}[H]
\centering
\includegraphics[width=\linewidth]{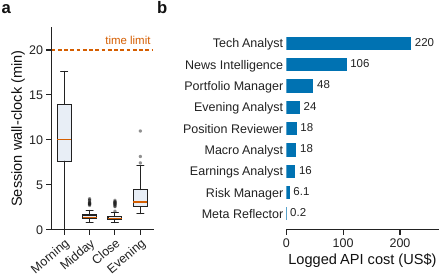}
\Description{Two panels. a, box plots of session wall-clock: morning sessions take about ten minutes at the median, midday and close about one to two minutes, evening about three; all lie below the 20-minute hard kill. b, horizontal bars of logged API cost per agent, led by the Technical Analyst at about 220 dollars and News Intelligence at about 106 dollars.}
\caption{(a)~Session wall-clock; dashed line, the session time limit. (b)~Logged API cost per agent over the deployment (a lower bound: per-call cost logging began on May~12).}
\label{fig:ops}
\end{figure}

Table~\ref{tab:sessions} lists the daily sessions; Figure~\ref{fig:ops} gives wall-clock and cost from the trace: the morning session, which runs four analysts, the Portfolio Manager and the Risk Manager, takes ten minutes at the median and always finished within the 20-minute session limit; half of the cost is the Technical Analyst scoring the whole universe each morning.

\begin{table}[H]
\caption{Daily sessions (US Eastern, Mon--Fri). ``LLM'' marks sessions that call language models.}
\label{tab:sessions}
\scriptsize
\setlength{\tabcolsep}{3pt}
\begin{tabular}{@{}llcp{3.55cm}@{}}
\toprule
Window & Session & LLM & Scope \\
\midrule
08:00--09:15 & earnings\_preprocess & \checkmark & analyse newly filed 10-Q/10-K \\
09:30--12:00 & morning & \checkmark & research, decide, risk-check, execute; the only session that may buy \\
09:30--16:00 & intra\_check (30') & -- & 3\% daily-loss circuit breaker \\
13:00--14:30 & midday & \checkmark & sell-only review (patient) \\
15:30--16:00 & close & \checkmark & sell-only review (act on hard triggers) \\
20:00--22:00 & evening & \checkmark & reconcile fills; grade trades and forecasts; write insights \\
quarter end & meta & \checkmark & self-portrait over 90 days; propose learnings; governed apply \\
\bottomrule
\end{tabular}
\end{table}

\section{Additional Evaluation Material}
\label{app:eval}
\releaseAppendix

\paragraph{Scoring and simulation.} Forward returns are close to close over 1, 5, 10 and 20 trading days and are market-adjusted by subtracting SPY's return over the same horizon. Means are taken first within each day and then across days, so that a day with many ratings counts once, and standard errors are computed on the daily means. The rank IC is the daily Spearman correlation between a signal and the market-adjusted forward return, on days with at least eight scored symbols. The policy-layer simulations of Table~\ref{tab:ablation} hold each recorded target from the decision day's close, cap a name at 20\% and the book at 100\%, and earn the T-bill ETF's return on idle cash. With system exits they apply the Technical Analyst's stop from the entry day or the three days before it (8\% below entry if none) and the Position Reviewer's recorded SELL, REDUCE (halving) and TRAIL\_STOP actions; with the Risk Manager, a rejected plan sets no targets and its recorded allocation edits and scale factors are applied. The analyst-following baselines put 5\% in every name rated buy (or high-conviction buy), up to 70\% of equity.

\begin{figure}[H]
\centering
\includegraphics[width=\linewidth]{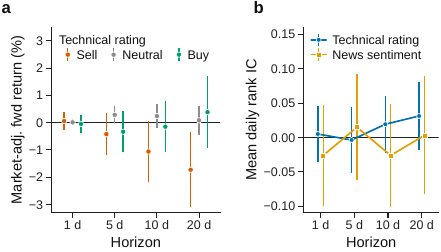}
\Description{Two panels. a, mean market-adjusted forward return after Technical Analyst sell, neutral and buy ratings at 1, 5, 10 and 20 days; only sell ratings at 10 and 20 days are clearly negative. b, mean daily rank information coefficient of the Technical rating and of News sentiment by horizon; all intervals include zero, the Technical IC rising to about 0.03 at 20 days.}
\caption{Analyst signal quality. (a)~Market-adjusted forward return after each Technical Analyst rating. (b)~Mean daily Spearman rank IC between signal and forward market-adjusted return. Mean and 95\% interval, day-clustered.}
\label{fig:signal}
\end{figure}

\paragraph{Analyst signals and forecasts (RQ1).} Figure~\ref{fig:signal} scores analyst outputs against market-adjusted forward returns.

\paragraph{Exits (RQ1).} Table~\ref{tab:exits} splits the \nRoundTrips{} broker round trips by exit mechanism.

\begin{table}[H]
\caption{Exit quality. Post-exit 20d: the sold symbol's 20-day return after exit minus SPY's (``correct'' if negative); Left: further gain to the end of the study.}
\label{tab:exits}
\scriptsize
\setlength{\tabcolsep}{2.4pt}
\begin{tabular}{@{}lrrrrrrr@{}}
\toprule
 & & & & & \multicolumn{2}{c}{Post-exit 20d} & \\
\cmidrule(lr){6-7}
Exit & Lots & Win & Realised & Hold (d) & Excess & Correct & Left (US\$k) \\
\midrule
Agent decision & 306 & 51\% & $+$0.45\% & 9.6 & $+$0.44\% & 42\% & 19.9 \\
Mechanical stop & 132 & 39\% & $+$2.29\% & 15.3 & $+$4.06\% & 45\% & 0.7 \\
\midrule
All & 438 & 47\% & $+$1.01\% & 11.4 & $+$1.27\% & 43\% & 20.6 \\
\bottomrule
\end{tabular}

\end{table}

\paragraph{Self-evolution (RQ3).} Table~\ref{tab:replay} details the Q2 replay, Table~\ref{tab:q3} the Q3 replay and Table~\ref{tab:feedback} the evening-loop correlations. The Q2 replay feeds recorded mornings through the Portfolio Manager prompt with and without the withheld learning; the Q3 replay runs frozen evening inputs through the Evening Analyst prompt before and after the commit, two samples each, where a trigger evening has a symbol classified as a missed value entry or trend timing on at least three of the last ten evenings.

\begin{table}[H]
\caption{Replay of the withheld Q2 learning on the Portfolio Manager. Flag.: flagged BUYs; Base: of which the baseline still proposed; Kept: within $\pm$1 point.}
\label{tab:replay}
\scriptsize
\setlength{\tabcolsep}{3pt}
\begin{tabular}{@{}lrrrrrrr@{}}
\toprule
 & & & \multicolumn{5}{c}{Learning vs.\ baseline} \\
\cmidrule(l){4-8}
Mornings & Flag. & Base & Kept & Red. & Incr. & Drop. & Added \\
\midrule
Wrong BUYs (10 d) & 15 & 10 & 8 & 1 & 1 & 0 & 1 \\
Control BUYs (5 d) & 14 & 12 & 8 & 3 & 1 & 0 & 0 \\
\midrule
\multicolumn{8}{@{}p{0.97\linewidth}@{}}{Summed weight of flagged BUYs, wrong mornings: 59.6\% $\to$ 63.5\%; control: 72.5\% $\to$ 67.3\%. Re-running the baseline itself moves a flagged weight by 1.8pp at the median ($n$=7 pairs). 35 LLM calls, US\$17.} \\
\bottomrule
\end{tabular}

\end{table}

\begin{table}[H]
\caption{Replay of the applied Q3 learning on the Evening Analyst. Recurring named: recurring missed symbols named in suggested actions; TAKE/WAIT, Blocker: replays with an explicit call or a named blocker.}
\label{tab:q3}
\scriptsize
\setlength{\tabcolsep}{3pt}
\begin{tabular}{@{}llrrrr@{}}
\toprule
Evenings & Prompt & Replays & Recurring named & TAKE/WAIT & Blocker \\
\midrule
Trigger (10) & base & 20 & 26/64 & 0/20 & 0/20 \\
 & + learning & 20 & 25/64 & 3/20 & 4/20 \\
\midrule
Control (4) & base & 8 & 0/2 & 0/8 & 0/8 \\
 & + learning & 8 & 0/2 & 0/8 & 0/8 \\
\bottomrule
\end{tabular}

\end{table}

\begin{table}[H]
\caption{Evening loop: Spearman correlations between trailing code-computed grades and next-morning behaviour, with raw and Holm-corrected $p$-values.}
\label{tab:feedback}
\scriptsize
\setlength{\tabcolsep}{3pt}
\begin{tabular}{@{}lrrrr@{}}
\toprule
Relation (Spearman) & $\rho$ & $p$ & $p_{\mathrm{Holm}}$ & $n$ \\
\midrule
New BUY weight $\sim$ trailing 10-evening forecast hit rate & $-$0.14 & 0.32 & 1.00 & 54 \\
Number of BUYs $\sim$ trailing forecast hit rate & $-$0.11 & 0.41 & 1.00 & 54 \\
New BUY weight $\sim$ RM cuts in last 5 sessions & $-$0.08 & 0.49 & 1.00 & 82 \\
New BUY weight $\sim$ trailing 45-day realised win rate & $+$0.02 & 0.89 & 1.00 & 80 \\
Book exposure $\sim$ trailing 45-day realised win rate & $+$0.22 & 0.05 & 0.32 & 80 \\
Book exposure $\sim$ trailing forecast hit rate & $+$0.34 & 0.01 & 0.08 & 54 \\
\bottomrule
\end{tabular}

\end{table}

\begin{figure}[H]
\centering
\includegraphics[width=\linewidth]{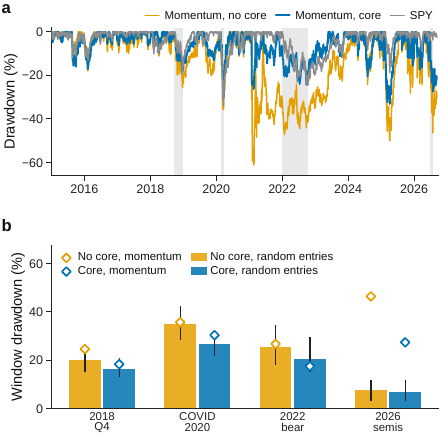}
\Description{Two panels. a, drawdown over 2015 to 2026 of a momentum portfolio without and with the core, and of SPY; with the core the drawdowns are shallower, especially in 2021 and 2022 where the no-core portfolio falls about 60 percent. b, drawdown within four crisis windows, 2018 Q4, COVID 2020, the 2022 bear market and the 2026 semiconductor sell-off: with random entries the median drawdown is lower with the core in every window; momentum entries show the same pattern, most strongly in 2026 where the core reduces the loss from about 46 to 27 percent.}
\caption{Deterministic core, 2015--2026. (a)~Drawdown of momentum entries without and with the core, and of SPY; shaded, crisis windows. (b)~Drawdown per window: median and 5--95\% range over 50 random seeds; diamonds, momentum.}
\label{fig:backtest}
\end{figure}

\paragraph{Containment and grounding (RQ4, RQ5).} Table~\ref{tab:backtest} and Figure~\ref{fig:backtest} give the 2015--2026 backtest of the deterministic core; Table~\ref{tab:grounding} gives the grounding audit by agent.

\paragraph{Backtest protocol.} Entries are rebalanced on the first trading day of each week to ten names at 9\% each: the top ten by 126-day return skipping the last five days (momentum), or ten names drawn uniformly (random, 50 seeds). Without the core, the book holds 10\% per name, fully invested, with no stop, cap or breaker. The core applies a stop at entry minus two ATR(14), triggered on the daily low and filled at the lower of the open and the stop; a 40\% GICS sector cap; at most 90\% net exposure; the cash-only rule; and a breaker that liquidates at the close after a day of $-3\%$ or worse and re-enters at the next open. Stopped names stay flat until the next rebalance. Trades cost 5\,bp per side, and idle cash earns a T-bill ETF.

\paragraph{Grounding protocol.} The audit samples up to 150 CoTs per agent and extracts every number in their reasoning fields, excluding small integers and years. A number is grounded if it occurs in the agent's input message, in any historical version of its prompt or in its own emitted fields, allowing rounding and truncation; the derived matcher also accepts risk/reward ratios, ATR multiples, percentage changes, weights and stop levels computed from those values, and arithmetic written out in the text. The placebo pairs each CoT with the input of another row of the same agent, half the sample apart in time; because two-digit numbers such as 25\% or 1.5 match by chance, the placebo is reported on numbers with three or more significant digits.

\begin{table}[H]
\caption{Backtest of the deterministic core alone, 2015--2026, on the 80 single stocks of the universe with LLM-free weekly entries (top 10 by 126-day momentum, or random over 50 seeds). Core: the rules of \S\ref{sec:core}; ablations remove one. CAGR and drawdowns in \%; 5\,bp per side. The universe is from 2026, so only comparisons between variants are meaningful.}
\label{tab:backtest}
\scriptsize
\setlength{\tabcolsep}{3.2pt}
\begin{tabular}{@{}lrrrrrrr@{}}
\toprule
 & & & & \multicolumn{4}{c}{Crisis drawdown} \\
\cmidrule(l){5-8}
Entries / variant & CAGR & Sharpe & Max DD & 2018 & 2020 & 2022 & 2026 \\
\midrule
\multicolumn{8}{@{}l}{\emph{Momentum entries}} \\
\quad No core & $+$61 & 1.15 & $-$61 & $-$25 & $-$36 & $-$27 & $-$46 \\
\quad Stops only & $+$58 & 1.12 & $-$61 & $-$25 & $-$29 & $-$27 & $-$51 \\
\quad Core $-$ stops & $+$40 & 1.26 & $-$36 & $-$20 & $-$36 & $-$21 & $-$27 \\
\quad Core $-$ breaker & $+$50 & 1.09 & $-$60 & $-$23 & $-$25 & $-$19 & $-$39 \\
\textbf{\quad Full core} & \textbf{$+$41} & \textbf{1.30} & \textbf{$-$33} & \textbf{$-$18} & \textbf{$-$30} & \textbf{$-$18} & \textbf{$-$27} \\
\multicolumn{8}{@{}l}{\emph{Random entries, median of 50 seeds}} \\
\quad No core & $+$17 & 0.81 & $-$37 & $-$20 & $-$35 & $-$26 & $-$8 \\
\textbf{\quad Full core} & \textbf{$+$14} & \textbf{0.82} & \textbf{$-$28} & \textbf{$-$17} & \textbf{$-$27} & \textbf{$-$21} & \textbf{$-$7} \\
\midrule
SPY & $+$14 & 0.82 & $-$34 & $-$19 & $-$34 & $-$24 & $-$3 \\
\bottomrule
\end{tabular}

\end{table}

\paragraph{Replay noise.} The deployed design runs twice per morning to measure re-run noise.

\begin{table}[H]
\caption{Number-grounding audit of recorded CoTs. Grounded: cited numbers found in the agent's input, prompt or own fields, including derived quantities; Strict: exact or rounded matches only; Placebo: strict grounding against another day's input.}
\label{tab:grounding}
\scriptsize
\setlength{\tabcolsep}{4pt}
\begin{tabular}{@{}lrrrrr@{}}
\toprule
Agent & Rows & Numbers & Grounded & Strict & Placebo \\
\midrule
Portfolio Manager & 133 & 7,827 & 99.69\% & 98.85\% & 41\% \\
Risk Manager & 75 & 979 & 99.59\% & 91.52\% & 18\% \\
Macro Analyst & 137 & 3,457 & 99.94\% & 99.28\% & 9\% \\
Technical Analyst & 150 & 233,310 & 99.94\% & 95.85\% & 23\% \\
Position Reviewer & 150 & 3,155 & 99.40\% & 98.07\% & 21\% \\
Evening Analyst & 105 & 5,366 & 99.27\% & 98.53\% & 17\% \\
\midrule
All & 750 & 254,094 & 99.91\% & 96.06\% & 22\% \\
\bottomrule
\end{tabular}

\end{table}

\section{Web Input, Taxonomies and Failure Handling}
\label{app:impl}
\paragraph{Web input.} RSS items are deduplicated by normalised URL and then by title-word Jaccard similarity above 0.7, their summaries are truncated to 300 characters, and at most 50 items enter the News agent's prompt. Filing text is parsed from HTML and capped at 30{,}000 characters. The Portfolio Manager receives at most three news items per symbol, each reduced to its conviction, sentiment and impact summary, with no headline. Every news state change is checked by code against the session's input and the prior state, and dropped if none of its keywords or symbols occurs there.

\paragraph{Evening taxonomy.} A \texttt{wrong} BUY carries one of nine root causes: greed or top chasing, macro warning ignored (which must cite the warning), herd buying, averaging down, broken thesis held, concentration, timing, systemic drawdown and tail event, together with a thesis trajectory (strengthening, intact, weakening, broken). A missed move is classified as a trend-timing miss, a theme blind spot, a fundamentals mispricing, a missed value entry, a noise rally or a risk-disciplined pass; the last two are not counted as misses. The Risk Manager labels each verdict with one of ten reason categories (for example \emph{oversized}, \emph{rr\_fail}, \emph{concentration}, \emph{event risk}, \emph{data degraded}), which L5 feeds back to the Portfolio Manager.

\paragraph{Failure handling.} Table~\ref{tab:failures} lists how the system contains each class of failure. The rule throughout is that a failure removes an action rather than adding one.

\begin{table}[H]
\caption{Failure classes and the system's response.}
\label{tab:failures}
\scriptsize
\setlength{\tabcolsep}{3pt}
\begin{tabular}{@{}p{0.36\linewidth}p{0.60\linewidth}@{}}
\toprule
Failure & Response \\
\midrule
Transient LLM error (429, 5xx) & up to 7 retries with backoff, then one call to a second vendor's model \\
Unparseable or invalid output & the agent's output is dropped; invalid list items are dropped singly \\
Research agent fails & recorded as degraded; two degraded sources raise a Risk Manager advisory \\
Portfolio Manager fails & session aborts without orders; the next 30-minute tick retries \\
Risk Manager fails & the plan is rejected (fail closed) \\
Position Reviewer fails & no exit actions; broker-held stops remain \\
Evening Analyst fails & P\&L is recorded; no grades are written \\
Crash between order and record & write-ahead record and reconciliation of orphans \\
Stale entry price ($>$5\% from market) & the BUY is skipped \\
Daily loss beyond 3\% & no new buys; the circuit breaker cancels entries and liquidates \\
Session overrun & killed after 20 minutes; at most one run per session per day \\
\bottomrule
\end{tabular}
\end{table}

\paragraph{Ablations.} With momentum entries the core's drawdown is 36\% without stops and 60\% without the breaker (33\% in full).

\end{document}